\documentclass[conference]{IEEEtran}
\IEEEoverridecommandlockouts
\usepackage{cite}
\usepackage{amsmath,amssymb,amsfonts}
\usepackage{algorithmic}
\usepackage{graphicx}
\usepackage{textcomp}
\usepackage{xcolor}
\usepackage{eucal}
\usepackage{esvect}
\usepackage{booktabs}
\usepackage{multirow}
\usepackage{hyperref}
\def\BibTeX{{\rm B\kern-.05em{\sc i\kern-.025em b}\kern-.08em
    T\kern-.1667em\lower.7ex\hbox{E}\kern-.125emX}}
\begin{document}

\title{CaRoBio: 3D Cable Routing with a Bio-inspired Gripper Fingernail}


\author{
\IEEEauthorblockN{1\textsuperscript{st} Jiahui Zuo}
\IEEEauthorblockA{\textit{Department of Electronic and Computer Engineering} \\
\textit{Hong Kong University of Science and Technology}\\
Hong Kong, China \\
jzuoai@connect.ust.hk}
\and
\IEEEauthorblockN{2\textsuperscript{nd} Boyang Zhang}
\IEEEauthorblockA{\textit{Department of Electronic and Computer Engineering} \\
\textit{Hong Kong University of Science and Technology}\\
Hong Kong, China \\
bzhangcd@connect.ust.hk}
\and
\IEEEauthorblockN{3\textsuperscript{rd} Fumin Zhang}
\IEEEauthorblockA{\textit{Department of Electronic and Computer Engineering} \\
\textit{Hong Kong University of Science and Technology}\\
Hong Kong, China \\
eefumin@ust.hk}
}

\maketitle 

\begin{abstract}
The manipulation of deformable linear flexures has a wide range of applications in industry, such as cable routing in automotive manufacturing and textile production. Cable routing, as a complex multi-stage robot manipulation scenario, is a challenging task for robot automation. Common parallel two-finger grippers have the risk of over-squeezing and over-tension when grasping and guiding cables. In this paper, a novel eagle-inspired fingernail is designed and mounted on the gripper fingers, which helps with cable grasping on planar surfaces and in-hand cable guiding operations. Then we present a single-grasp end-to-end 3D cable routing framework utilizing the proposed fingernails, instead of the common pick-and-place strategy. Continuous control is achieved to efficiently manipulate cables through vision-based state estimation of task configurations and offline trajectory planning based on motion primitives. We evaluate the effectiveness of the proposed framework with a variety of cables and channel slots, significantly outperforming the pick-and-place manipulation process under equivalent perceptual conditions. Our reconfigurable task setting and the proposed framework provide a reference for future cable routing manipulations in 3D space.
\end{abstract}


\section{Introduction}


Robot manipulation on deformable linear objects (DLOs) is rapidly developing recently, with applications in industrial and everyday scenarios.
\begin{figure}[htbp]
    \centering
    \includegraphics[width=1.0\linewidth]{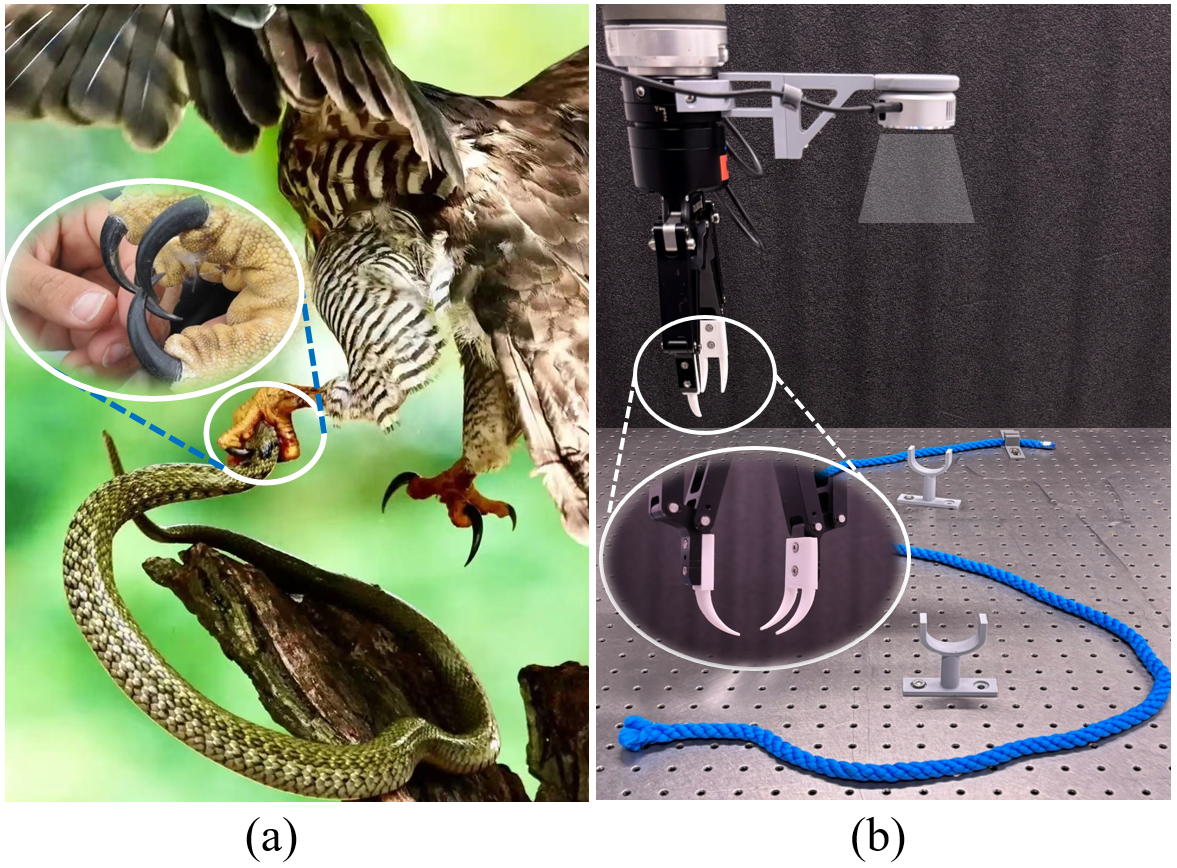}
    \caption{
    Eagle-inspired robotic system design: From natural predation to cable manipulation.
    (a) Eagle catching a snake and zoomed-in eagle claw.
    (b) Robotic cable routing system and zoomed-in eagle-inspired fingernails.}
    \label{cover}
\end{figure}
Related human daily tasks surrounding DLOs, including tying shoelaces, organizing USB cables, and hanging clotheslines, still pose significant challenges to robots.
These difficulties arise from DLOs' high-dimensional configuration space, strongly nonlinear dynamic behaviors, and behavioral uncertainties caused by variations in material properties and geometric parameters (e.g., length and diameter)\cite{9732674}. Recent works have explored various approaches including specialized end-effectors \cite{9981313,9571044}, multi-modal sensing \cite{chen2023contact,wilson2023cable}, and hierarchical planning \cite{luo2307multi,jin2022robotic}. 

However, automated DLO manipulation remains predominantly confined to laboratory settings. This is particularly evident in industrial cable routing tasks that still rely heavily on manual labor,  despite widespread automation in other assembly processes.

Functional specialization approaches include integrated multi-modal grippers for flat cables \cite{9981313}, pneumatic soft grippers for high-voltage lines \cite{9347270}, and three-mode grippers for noodle alignment \cite{10417506}. In addition to humans, many animals in the natural world possess the ability to manipulate flexible linear objects. For instance, eagles use their sharp claws to capture agile snakes, and storks capture elongated eels with their serrated beaks. Bio-inspired solutions show particular promise, with gecko-inspired adhesion layers reducing squeezing forces \cite{7139505} and V-shaped dexterous hands enabling in-hand cable following \cite{Yu2024InHandFO}. However, most designs either require complex mechanisms \cite{9981313} or prove inadequate for pickup tasks on horizontal tables \cite{9816981}. 
Current DLO routing operations mainly rely on conventional industrial parallel grippers, which often risk cable slippage, over-squeezing, or over-tension, and limit manipulation to 2D planes.

This limitation motivates our biological investigation of eagle predation strategies, in which curved claws enable stable and secure snake grasping, as shown in Fig. \ref{cover}(a).
By adjusting claw angles and grasping forces, they counteract reactive forces generated by the snake's flexible undulations. Their curved fingernails effectively hook the snake's body to prevent it from slipping away - a capability notably absent in conventional parallel grippers. 
Inspired by this biological insight, we wondered if it would be possible to enhance the ability to manipulate DLOs by fingertip reconstruction \cite{9197396} that simply equips industrial parallel gripping jaws with eagle-inspired fingernails.



We set up a simplified reconfigurable 3D workspace for cable routing tasks as shown in Fig. \ref{cover}(b). The cable is initially placed on a planar optical table with one fixed end, and different specifications of slots randomly stand on the table. 
The robot arm requires achieving accurate cable pickup from the planar surface without extreme squeezing. Then it can manipulate the cable through several action sequences to guarantee that it passes the slots.
To simplify the problem, we formulate the following task assumptions: 1) the cable stays within the RGBD camera field of view throughout the manipulation; 2) the cable is distinguishable by contrast of color with the background and there is no heavy occlusion. 3) cables are required to be inserted into different slots from the fixed end to the loose end.

Algorithmic innovations present similar trade-offs between complexity and practicality. Hierarchical frameworks combining visual perception with motion primitives \cite{jin2022robotic,waltersson2022planning} achieve planar cable routing but typically require dual-arm coordination \cite{chen2023contact} or continuous sensor feedback \cite{suberkrub2022feel}. Tactile-visual fusion methods \cite{de2018integration,wilson2023cable} improve insertion accuracy at the cost of increased system complexity. Our work addresses these limitations through a single-grasp paradigm that leverages geometric constraints rather than continuous sensing, enabled by our claw-like fingernail design.  
Specifically, we propose a single-grasp end-to-end framework for 3D cable routing. Initialized with a single RGBD observation capturing both cable and support slots, the perception model first detects slot poses through depth threshold filtering complemented by PCA-based orientation estimation. Meanwhile, cable geometry is extracted via YOLOv8 instance segmentation followed by morphological refinement to obtain precise cable masks. 
A cable preprocessing step is designed to adjust the cable to an idealized configuration for subsequent planning. The task configuration (slot position, orientation and cable nodes) is mapped to a trajectory through four motion primitives. With a multi-weight voting strategy, the optimal grasping node is determined. Sequentially, the cable is manipulated with a sequence of parameterized motion primitives iteratively which grasp only once and maintain continuous cable contact.

The main contributions presented in this work can be summarized as follows:

\indent 1) We propose a \textbf{Bio-inspired Fingernail} design capable of two grasping modes (slack grasping, tight grasping) for DLO manipulation. With Force-Stroke Coupling mechanism, this design enhances grasping adaptability and operational stability without tactile feedback, while reducing cable extrusion deformation. 

\indent 2) We present a \textbf{Single-Grasp End-to-End} framework instead of common pick-and-place strategy. After perception, a node-based cable preprocessing method and multi-weight voting grasp decision are used to handle various cable routing task configurations. The configuration datas are translated into a executable cable routing trajectory through four motion primitives. The framework has proven to be an effective solution during experiments.
\section{Bio-inspired Fingernail Design}
\label{sec4}

\subsection{Appearance and Features} 
\label{sec4_1}

\begin{figure}[htbp]
    \centering
    \includegraphics[width=1.0\linewidth]{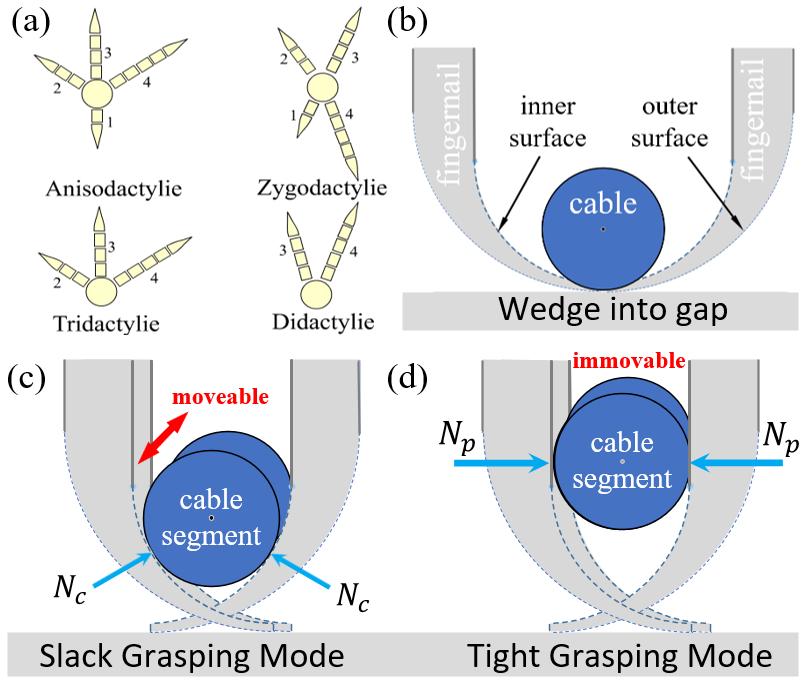}
    \caption{ 
    Bio-inspired grasping mechanisms.
    (a) Types of bird feet.
    (b) Front view of grasping by wedging into the gap.
    (c) Slack grasping mode.
    (d) Tight grasping mode.}
    
    \label{modes}
\end{figure}

Traditional parallel grippers use planar friction to handle cables but struggle with cable slippage, over-squeezing, or over-stretching. Inadequate friction demands excessive squeezing forces, risking insulation damage of soft cables. Limited contact area causes uneven pressure and stress concentrations, increasing failure risks. 
During dynamic tasks (e.g., guiding), tangential force fluctuations induce slippage, limiting precise 3D cable shaping and positioning.

We investigate different types of bird feet shown in Fig. \ref{modes}(a). Bird feet are well adapted to the life they lead. Most birds have four toes, inherited  from their theropod dinosaur ancestors. These four toes are arranged into four main patterns: Anisodactylie, Syndactylie, Tridactylie, and Didactylie. Anisodactylie is the most common toe arrangement in birds, with three toes forward and one toe back. It is the basic pattern of hunting birds such as eagles and falcons. Our design takes a page from the eagle's Anisodactylie pattern, simplifying it to two fingernails forward and one fingernail back shown in Fig. \ref{cover}(b).
The proposed bio-inspired fingernail features dual-arc contact surfaces optimized for cable manipulation:
\subsubsection{Shape of fingernail} 

The inner contact surface is a quarter-circle arc of radius $R_i$, tangent to the gripper jaws for seamless contact during closure. The outer contact surface is a larger-radius arc ($R_o > R_i$) with its center directly above the inner arc's center. This vertical collinear alignment keeps the fingernail tip nearly horizontal, allowing it to wedge into narrow gaps between cable and surface and adapt to different cable diameters.

\subsubsection{Bidirectional Engagement} Three fingernails are arranged in a left-right interleaved configuration, driven by gripper opening/closing to bidirectionally wedge into the gap between the cable and surface, as shown in Fig. \ref{modes}(b). Balanced contact forces in lateral directions reduce cable lateral displacement and torsion, further enhancing the grasp stability.

\subsection{Mathematical Modeling of Grasping}
\label{sec4_2}

\begin{figure}[h]
    \centering
    \includegraphics[width=1.0\linewidth]{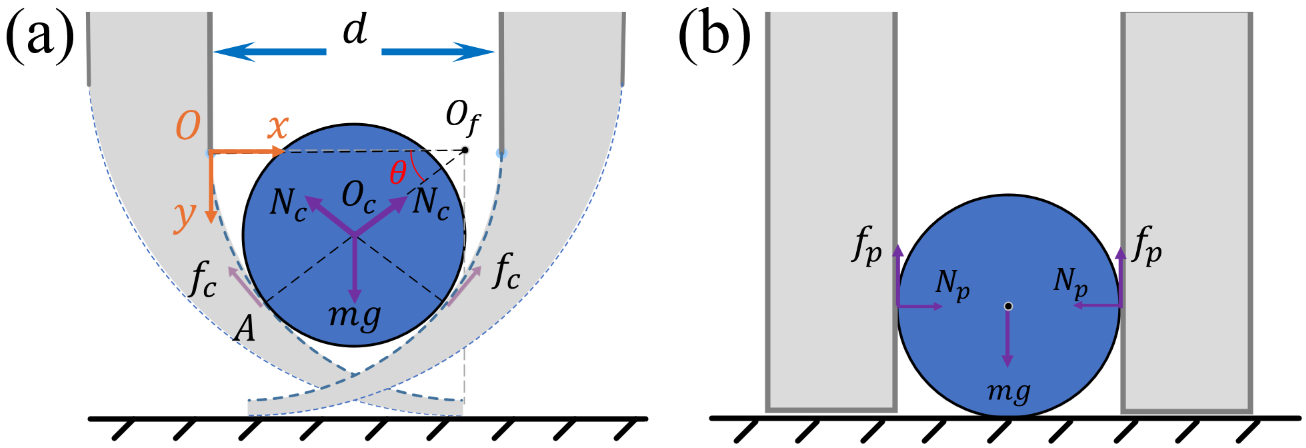}
    \caption{ 
    Mathematical model of different grasping method. 
    (a) Mathematical model of the bio-inspired fingernail. (b) Mathematical model of the parallel gripper. }
    \label{math_model}
\end{figure}

As depicted in Fig. \ref{math_model}(a), we establish a coordinate system with origin $O$ at the tangent point between the fingernail's inner contact surface and the parallel gripper jaws. From symmetry considerations, we derive:  
\begin{equation}
x_{O_{c}} = \frac{d}{2} \label{eq:x}
\end{equation}
where $d$ denotes the gripper’s stroke and \( x_{O_{c}} \) is the horizontal coordinate of the cable cross-section’s centroid $O_c$.

The coordinates of contact point $A(x_A, y_A)$ are determined through geometric analysis of the tangential contact condition between the cable's cross-section and the fingernail's internal contact surface. 
Let \(\mathcal{C}_f\) (fingernail) and \(\mathcal{C}_c\) (cable cross-section) denote two circles with centers \(O_f(x_{O_f}, 0)\) and \(O_c(x_{O_c}, y_{O_c})\), and radii \(R_f\) and \(R_c\), respectively. The following conditions hold:
\begin{equation}
\mathcal{C}_f : (x_A - x_{O_f})^2 + y_A^2 = R_f^2 \label{eq:fingernail}
\end{equation}
\begin{equation}
\mathcal{C}_c : (x_A - x_{O_c})^2 + (y_A - y_{O_c})^2 = R_c^2 \label{eq:cable}
\end{equation}
\begin{equation}
(x_{O_c} - x_{O_f})^2 + y_{O_c}^2 = (R_f - R_c)^2 \label{eq:tangency}
\end{equation}

This yields the angle $\theta$, which represents the inclination of the cable contact force $N_c$ relative to the horizontal direction.
\begin{equation}
\theta = \arctan \left( \frac{y_{A}}{x_{O{f}} - x_A} \right) \label{eq:theta}
\end{equation}

With static equilibrium equation (neglecting static friction), we can calculate the contact force $N_c$ exerted by the fingernail on the cable.
\begin{equation}
2N_c\sin{\theta} = mg \label{eq:fingernail}
\end{equation}
Based on the lifted cable segment length $L$, the corresponding mass can be determined as $m = \rho_l L$, where $\rho_l$ is the linear density. 
From Eqs.~(\ref{eq:x})--(\ref{eq:fingernail}), we conclude that the contact force $N_c$ is directly coupled to the jaw stroke $d$, yielding the functional relationship:
\begin{equation}
N_c = \mathcal{F}(d) = \frac{\rho_lLg(R_f - R_c)}{2((R_f - R_c)^2 - (R_f - \frac{d}{2})^2)^{1/2}}
\label{eq:FSC}
\end{equation}
By adjusting the gripper’s opening distance \(d\), the magnitude of the contact force \(N_c\) can be actively controlled, enabling small cable extrusion pressure and flexible manipulation.

For 2-finger parallel gripper in Fig. \ref{math_model}(b), we can estimate minimum contact force $N_p$ with
\begin{equation}
2\mu N_p = mg \label{eq:parallel_force}
\end{equation}
Normally, the actual contact force would be larger than $N_p$ for parallel gripper to achieve a stable grasp. And the cable is easy to be over-squeezed and damaged in the absence of tactile sensor feedback.

A comparison between the bio-inspired fingernail (Eq. \ref{eq:fingernail}) and the parallel gripper (Eq. \ref{eq:parallel_force}) contact force solutions reveals that when $\sin(\theta) > \mu$ (where $\mu$ is the coefficient of friction), the fingernail enables non-slip grasping at reduced contact forces by adjusting the gripper stroke $d$, which modulates both the contact point and applied force.

\subsection{Enhanced DLO Manipulation Ability}
\label{sec4_3}
The tunable gripper open distance $d$ and cable position adaptivity based on bio-inspired fingernail enable two adaptive grasping modes:
\begin{itemize}
    \item \textbf{Slack Grasping Mode (SGM):} With gripper opening distance \( d > D_c \) in Fig. \ref{modes}(c) (where \( D_c \) is the cable diameter), this low-force mode provides gentle radial constraint through the fingernail curvature, while allowing axial cable adjustments.(e.g., guiding or offsetting cables).
    
    \item \textbf{Tight Grasping Mode (TGM):} With gripper stroke \( d < D_c \) in Fig. \ref{modes}(d), this high-stability mode can enhance normal forces for secure cable inserting during interaction with slots.
\end{itemize}

This dual-mode capability facilitates seamless transitions between spatial positioning and manipulation, critical for multi-stage routing processes.



 

\section{Single-Grasp End-to-End Framework}
\label{sec4}

\begin{figure*}[htbp]
    \centering
    \includegraphics[width=0.9\linewidth]{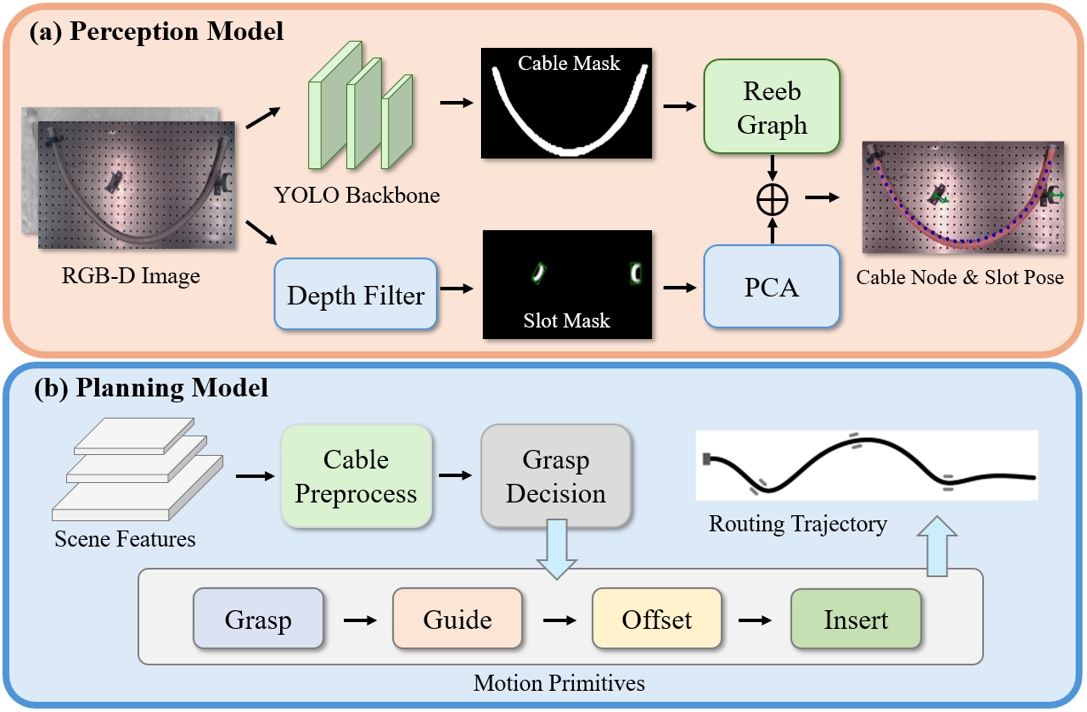}
    \caption{ \textbf{Pipeline overview:} We propose an single-grasp end-to-end framework for 3D cable routing. (a) Our perception model takes in one RGBD image of slots with initial cable to output the task configuration. We use depth filtering and Principle Component Analysis (PCA) to detect the position and orientation of the slots, and YOLOv8 to determine cable segmentation. (b) In planning model, a cable preprocessing step is designed to adjust the cable to a ideal state. With a multi-weight voting strategy, the optimal grasping node is determined. The scene features (slot positions, orientations and cable nodes) is mapped to a trajectory with four motion primitives. Sequentially, the cable is manipulated with a sequence of parameterized motion primitives iteratively which only grasp once.}
    \label{pipeline}
\end{figure*}

\label{sec4_1}

\subsection{Vision-based Task Configuration Perception}
An Intel RealSense Depth Camera L515 is mounted on the UR10 robotic wrist to simultaneously capture RGB-D images of both cables and slots, as shown in Fig. \ref{cover}(b). The perception model takes in RGB-D data from the depth camera overlooking the task plane and outputs the task configuration, as shown in Fig. \ref{pipeline}(a).

\textbf{Slot Pose Estimation:} For slot localization, a depth-filter-based masking algorithm is designed to extract slot regions. 
The slot's 3D pose (position and orientation) is determined by applying Principal Component Analysis (PCA) to its point cloud, extracting the eigenvector of the covariance matrix's largest eigenvalue as the major axis direction and combining it with the centroid position.
The central positions of the slots are denoted as
\(  {\mathcal{S}} = [ {S}_1,  {S}_2, \ldots,  {S}_K] \in \mathbb{R}^{K \times 3} \) and the orientations of the slots are denoted as \(  {\mathcal{\mathbf{O}}} = [ {\mathbf{O}}_1,  {\mathbf{O}}_2, \ldots,  {\mathbf{O}}_K] \in \mathbb{R}^{K \times 3} \)
where \(K\) is the number of slots. 

\textbf{Cable State Estimation \& Representation:} For cable segmentation, the YOLOv8 instance segmentation model \cite{10533619} is employed to achieve pixel-level cable mask extraction, and the Gaussian filter is used to smooth the depth image. The cable mask extracted from the color image and the filtered depth image was registered to generate \(3D\) cable point clouds. 
The generated point clouds are further refined by Euclidean clustering to eliminate noise points while preserving topological integrity. 
The positions of cable point clouds are denoted as 
\( \mathcal{P} = [P_1, P_2, \ldots, P_M] \in \mathbb{R}^{M \times 3} \).

With Reeb Graph algorithm \cite{6630714}, the cable point clouds are abstracted into a topologically connected backbone node chain. This representation facilitates downstream tasks with explicit spatial relationships. The positions of cable nodes are denoted as 
$  {\mathcal{N}} = [  {N}_1,  {N}_2, \dots,  {N}_Q ] \in \mathbb{R}^{Q \times 3} $
where \(Q\) is the predefined number of nodes. 
And the grasping orientation $\mathbf{V}_{N_{i}}$ of the node $ {N}_{i}$ can be represented by adjacent nodes 
\( \mathbf{V}_{N_{i}} =  {N}_{i+1} -  {N}_{i-1} \).
The cable status is defined as a set of 2-tuples \( \{ ( {N}_1, \mathbf{V}_{N_{1}}), ( {N}_2, \mathbf{V}_{N_{2}}),..., ( {N}_Q, \mathbf{V}_{N_{Q}}) \} \).

\label{sec4_2}

\subsection{Cable Preprocessing}
During our experiments, we observed that cables placed too close to slot bases often become entangled with the slots or bend excessively during 3D manipulation. This also makes it difficult for the gripper to reach optimal grasp points.

Existing planar cable routing research often relies on idealized initial conditions, such as sufficient cable-fixture distance and collision-free robotic sequences \cite{jin2022robotic}. However, these assumptions are too restrictive for spatial routing applications. Therefore, a preprocessing strategy is needed to adjust cable poses for smoother downstream manipulation.



\begin{figure}[htbp]
    \centering
    \includegraphics[width=1.0\linewidth]{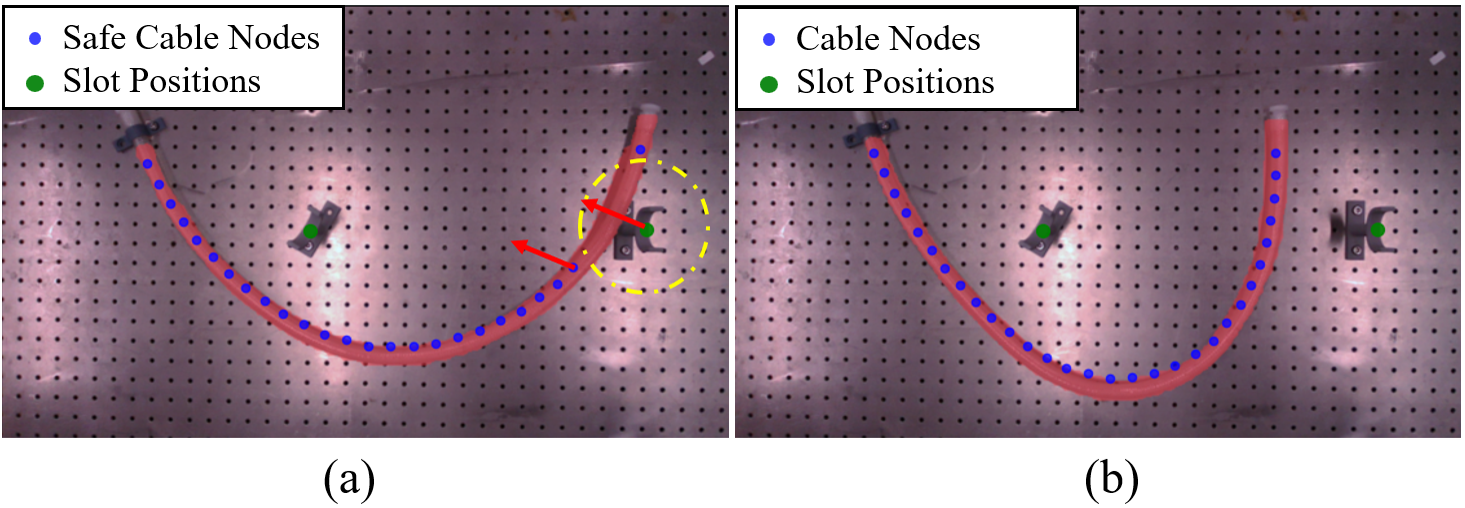}
    \caption{ Adjust cable position to avoid entanglement. (Yellow dashed circle: collision circle; Green square: closest cable node to the slot; Red arrow: adjustment vector).
    (a) Initial cable state.
    (b) cable state after adjustment.
     }
    \label{cable_node}
\end{figure}

\subsubsection{Set Collision Circle}
A cable safe node exclusion zone is defined as a circle (yellow dashed circle in Fig. \ref{cable_node}(a)) centered at the slot center \(  {S_j} \). The center radius \( R_c = R_s + R_g \), where \( R_s \) and \( R_g \) denote the slot's maximum radius and the gripper's outer profile radius respectively. 
Node $ {N}_{i}$ closest to \(  {S_j} \) is identified. An adjustment vector \( \mathbf{V}_a \) (red arrows in Fig. \ref{cable_node}(a)) is computed from the slot center to the closest cable node $ {N}_{i}$ , with magnitude \( R_c \), to adjust the cable away from collision-prone areas.

\subsubsection{Cable Adjustment} 
According to Motion Coherence Theory\cite{590011}, the points close to one another tend to move coherently. 
We assume a parallel motion between adjacent nodes, and thus, the nearest node \( {N}_k \) outside the exclusion zone is selected as the adjust point. The cable is translated along \( \mathbf{V}_a \) to maintain a safe distance from \(  {S_j} \). Iterate this step several times until all nodes are outside the collision circles.
After multiple collision avoidance adjustments, the cable has an ideal initial state where most nodes are outside the radius of the collision circle in Fig. \ref{cable_node}(b). 

\subsection{Grasping Point Selection Strategy}
\begin{figure}[htbp]
    \centering    \includegraphics[width=0.9\linewidth]{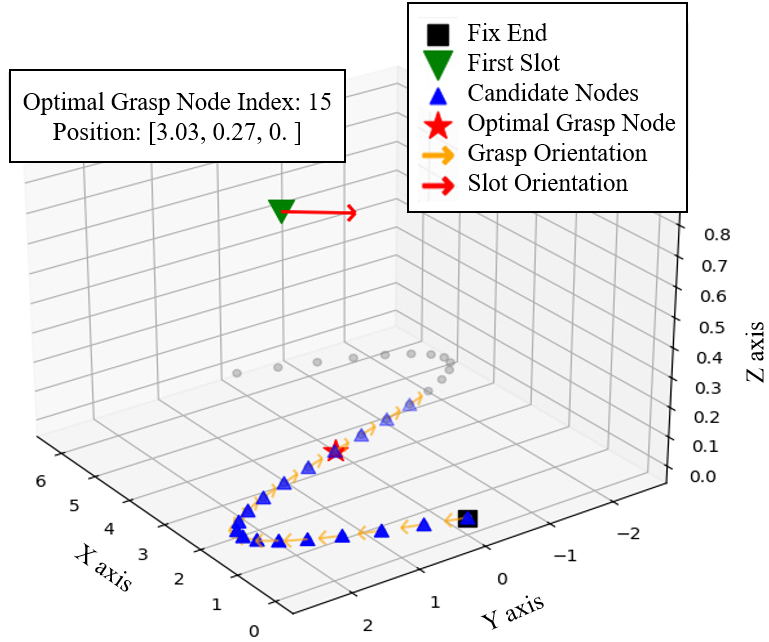}
    \caption{Cable grasping point selection result. The strategy gets candidates nodes and decides optimal grasping node based on distance factor and torsion angle factor.}
    \label{grasp_node}
\end{figure}

\label{Grasping}
With single-grasp manipulation planning from the fixed end to the moving end in \ref{sec3_3}, we only need to decide the optimal grasping node near the first slot \({S}_1\) to initiate the cable routing process (Fig. \ref{grasp_node}). 
Given the total number of cable nodes \( n \) and the cable length \( L_{\text{total}} \), the segment length \( L_i \) from the fixed end to node \( N_i \) is calculated as:  
\begin{equation}
    L_i = \frac{L_{\text{total}}}{n} \cdot i \quad (i = 1, 2, \dots, n).
\end{equation}  
To prevent excessive cable slack in subsequent cable guiding motion, only nodes \( N_i \) satisfying the geometric constraint $ L_i < D_{\text{fixed}\to S_1} $ are considered as valid candidates, where \( D_{\text{fixed}\to S_1} \) denotes the distance between the fixed end and the first slot \( S_1 \). 

\subsubsection{Factor Normalization}
Given node candidates $ {\mathcal{N}} = [  {N}_1,  {N}_2, \dots,  {N}_C ] \in \mathbb{R}^{C \times 3}$
where \(C\) is the number of candidate nodes, we mainly consider two factors: the moving distance factor ($D$) and the torsion angle factor ($\Theta$), representing the energy consumption of the robotic arm and the cable twisting damage respectively.

For distance factor, we compute the  Euclidean distance between candidate point $N_i$ and the first slot center $S_1$: 
\begin{equation}
D(N_i) = \|N_i - S_1\|
\end{equation}
and normalize it as: 
\begin{equation}
    D_{\text{norm}}(N_i) = \frac{D_{\text{max}} - D(N_i)}{D_{\text{max}} - D_{\text{min}}}
\end{equation}
Where \( D_{min} \) and \( D_{max} \) are the minimum and maximum node-slot distances.

For the torsion angle factor, we calculate the angle between grasping direction $\mathbf{V}_{N_{i}}$ at $N_i$ and slot orientation vector $\mathbf{V}_s$:
\begin{equation}
    \Theta(N_i) = \arccos\left(\frac{\mathbf{V}_{N_{i}} \cdot \mathbf{V}_s}{\|\mathbf{V}_{N_{i}}\| \|\mathbf{V}_s\|}\right)
\end{equation}
and normalize it as: 
\begin{equation}
    \Theta_{\text{norm}}(N_i) = 1 - \frac{\Theta(N_i)}{\pi/2}
\end{equation}
to ensure linear score decay for \( \Theta \in [0,90^\circ] \).

\begin{figure*}[htbp]
    \centering
    \includegraphics[width=1.0\linewidth]{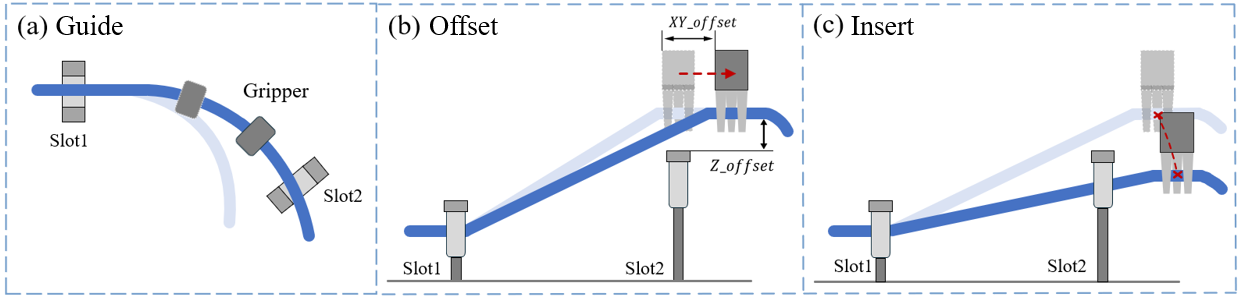}
    \caption{ Parameterized motion primitives.
     (a) Grasp motion primitive. (b) Offset motion primitive. (c) Insert motion primitive.}
    \label{motion}
\end{figure*}

\subsubsection{Multi-Weight Voting Strategy}
To balance distance and angular constraints, we propose a frequency-based voting mechanism to cover diverse scenarios by exploring weight configurations from pure distance-first ($w_D=1$) to pure angle-first ($w_\theta=1$) strategies.

First, we generate 50 linearly spaced weight pairs $(w_D^{(k)}, w_\theta^{(k)})$ from $(1, 0)$ to $(0, 1)$:
\begin{equation}
    w_D^{(k)} = 1 - 0.02(k-1), \quad w_\theta^{(k)} = 0.02(k-1), \quad k = 1, 2, \dots, 50
\end{equation}
For each weight pair $(w_D^{(k)}, w_\theta^{(k)})$, we compute the score for candidate $N_i$:
\begin{equation}
    \text{Score}_k(N_i) = w_D^{(k)} \cdot D_{\text{norm}}(N_i) + w_\theta^{(k)} \cdot \Theta_{\text{norm}}(N_i)
\end{equation}
The highest-scoring node $N_k^*$ for each weight pair is selected:
\begin{equation}
    N_k^* = \arg\max_{N_i \in \mathcal{N}} \text{Score}_k(N_i)
\end{equation}
Count frequency of each candidate in the 50 selected nodes $\{N_1^*, \dots, N_{50}^*\}$. The final grasp node $N_{\text{final}}^*$ is determined by:
\begin{equation}
    N_{\text{final}}^* = \arg\max_{N_i \in \mathcal{N}} \text{Freq}(N_i)
\end{equation}
If multiple nodes share the highest frequency, we prioritize the one with the highest balanced score:
\begin{equation}
    \text{Score}_{\text{base}}(N_i) = 0.5 \cdot D_{\text{norm}}(N_i) + 0.5 \cdot \Theta_{\text{norm}}(N_i)
\end{equation}

\subsection{Single-Grasp Manipulation Planning}

\label{sec3_3}
With the detected task configuration and two predefined grasping modes, a sequence of motion primitives is generated to execute cable routing:

\textbf{Grasp:} Optimal grasping point \( N_{\text{final}}^*  \) is selected based on node-slot distance and torsion angle from the grasping point selection strategy mentioned in \ref{Grasping}. And the gripper grasps the optimal cable node in slack grasping mode.


\textbf{Offset:} 
Due to the entity collision, the starting point and end point of the cable guiding trajectory cannot be strictly the slot center position, but need a certain offset. While maintaining slack cable grasping, the gripper adjusts the grasping point position through a translational motion along the slot axis or z direction, as shown in Fig. \ref{motion}(b). The horizontal offset vector along the slot axis at $S_j$ is defined as:
\begin{equation}
    \mathbf{d}^{\text{h}}_{S_{j}} = (\frac{w_g}{2} + \frac{w_s}{2} + \delta) {\mathbf{O}}_{j}
\end{equation} 
and the vertical offset vector is defined as:
\begin{equation}
    \mathbf{d}^{\text{v}}_{S_{j}} = (\frac{h_g}{2} + \frac{h_s}{2} + \delta)\mathbf{e_z}
\end{equation}
with the unit vector $\mathbf{e_z}$ in the $z$ direction.
(\( w_g \): gripper width; \( w_s \): slot width; \( w_g \): fingernail height; \( w_s \): slot height; \( \delta \): redundancy tolerance).

\textbf{Guide:} After grasping/inserting, SGM is restored to release cable stress. For the cable segment between \(  {S}_{j-1} \) and \(  {S}_j\), the guiding motion starts from point $G_{j-1} =  {S}_{j-1} + \mathbf{d}^{\text{h}}_{S_{j}} $ to point $G_{j} =  {S}_{j} + \mathbf{d}^{\text{v}}_{S_{j}} + \mathbf{d}^{\text{h}}_{S_{j}}$ while dynamically guiding the in-hand cable to the destination aligned with the slot axis, as shown in Fig. \ref{motion}(a). This avoids the over-tight/over-loose problem of the pick-and-place strategy \cite{jin2022robotic}, ensuring continuous stability in three-dimensional manipulation.

\textbf{Insert:} TGM is activated for cable inserting with a larger contact force at a constant rate \( v_{\text{insert}} \). When inserting the cable to the slot \(  {S}_{j} \), the insertion motion follows a circular arc trajectory centered at the previous starting point \( G_{j-1} \), as shown in Fig. \ref{motion}(c). 


As shown in Fig .\ref{pipeline}(b), the Single-Grasp manipulation sequences are defined as:

\label{sec4_5}
\begin{itemize}
    \item \textbf{Phase 1 (Preprocessing \& Grasp):} Adjust the cable to a collision-free initial state. Select optimal grasp node \( N_{\text{final}}^*  \) based on grasping point selection strategy.
    
    \item \textbf{Phase 2 (Guiding \& Offset):} Initiate SGM. The robotic arm moves toward the next slot with offset while guiding the in-hand cable to the next slot.
    
    \item \textbf{Phase 3 (Insertion):} Switch to TGM. Conduct cable insertion in a circle motion.
    
    \item \textbf{Phase 4 (Task Iteration):} Reset to SGM. Iterate Phase 2 and 3 until task completion.
\end{itemize}

\begin{figure}[htbp]
    \centering
    \includegraphics[width=0.9\linewidth]{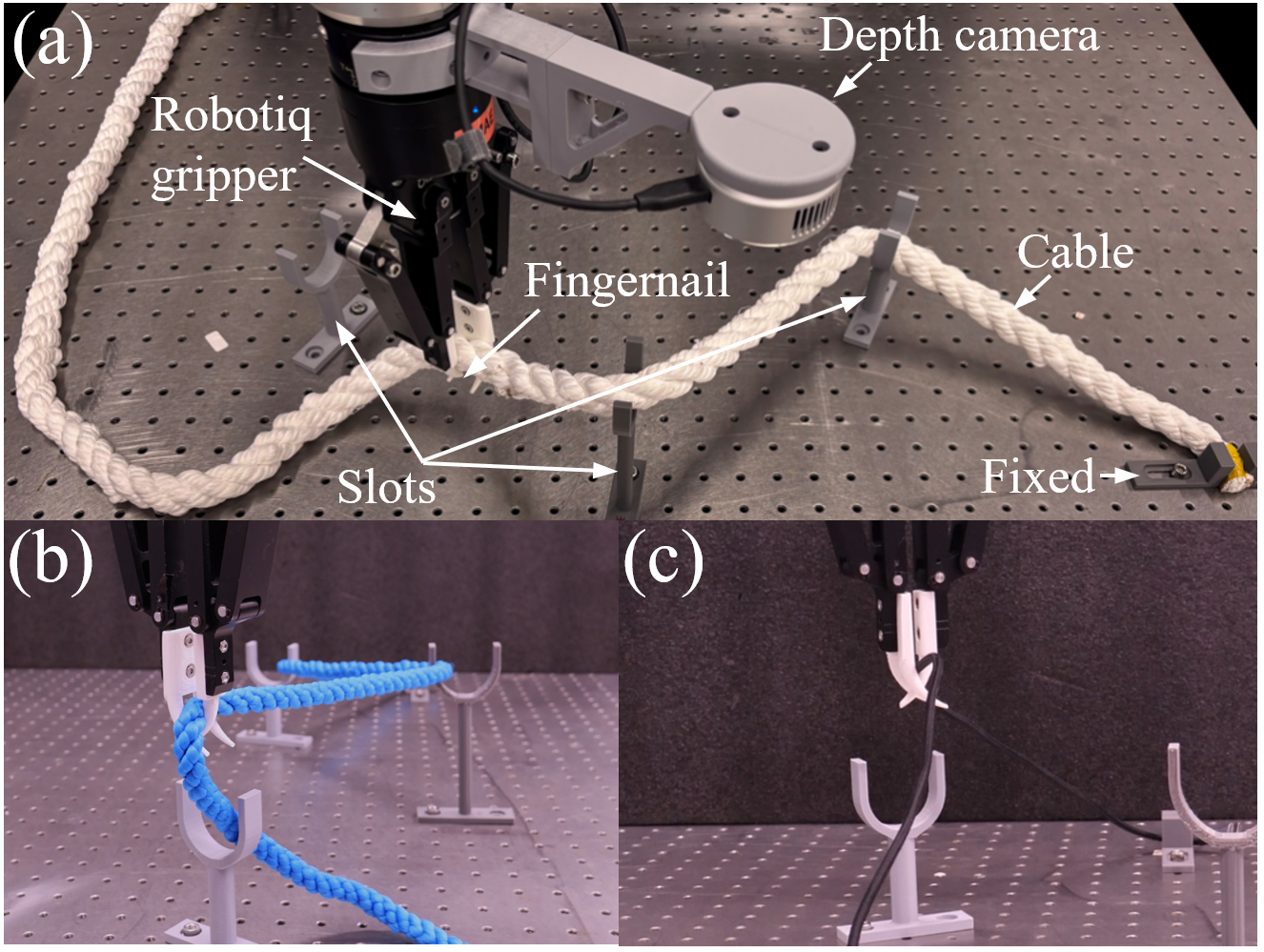}
    \caption{
    Experimental setup and failure samples.
    (a) Experimental setup.
    (b) Cable got folded.
    (c) Cable unaligned with slot.
    }
    \label{cable_rout_fail}
\end{figure}

\section{Experiments and Results}
\label{sec5}
Our system comprises a 6-DoF UR10 robot manipulator, an Intel RealSense L515 RGB-D camera, a Robotiq 2F-140 gripper, two bio-inspired fingernails, several cable slots, and a non-stretchable deformable cable as shown in Fig. \ref{cable_rout_fail} (a).

\label{sec5_1}
\begin{figure}[htbp]
    \centering
    \includegraphics[width=1.0\linewidth]{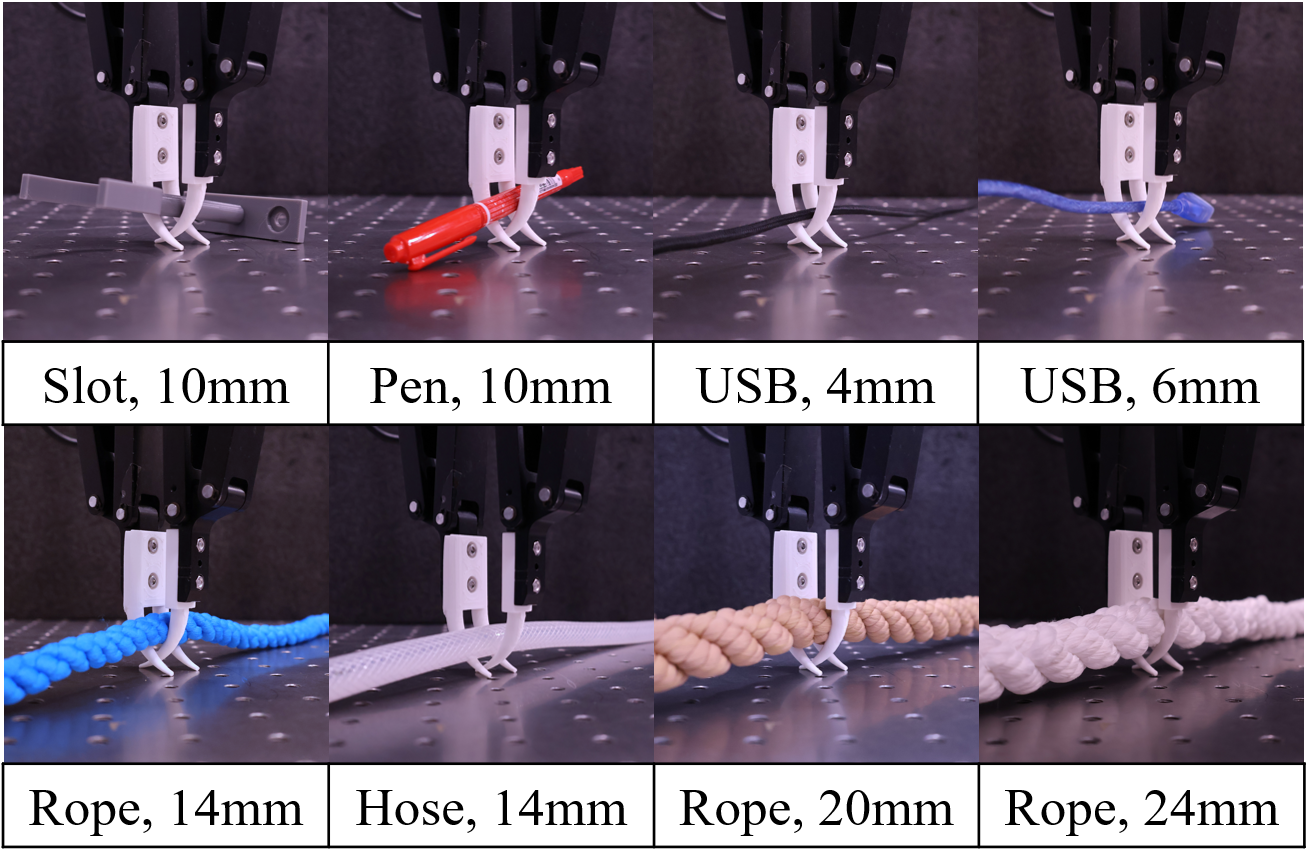}
    \caption{
    Grasping experiments with linear objects of different rigidities and diameters on the optical table.
    }
    \label{grasp}
\end{figure}

\begin{figure*}[htbp]
    \centering
    \includegraphics[width=1.0\linewidth]{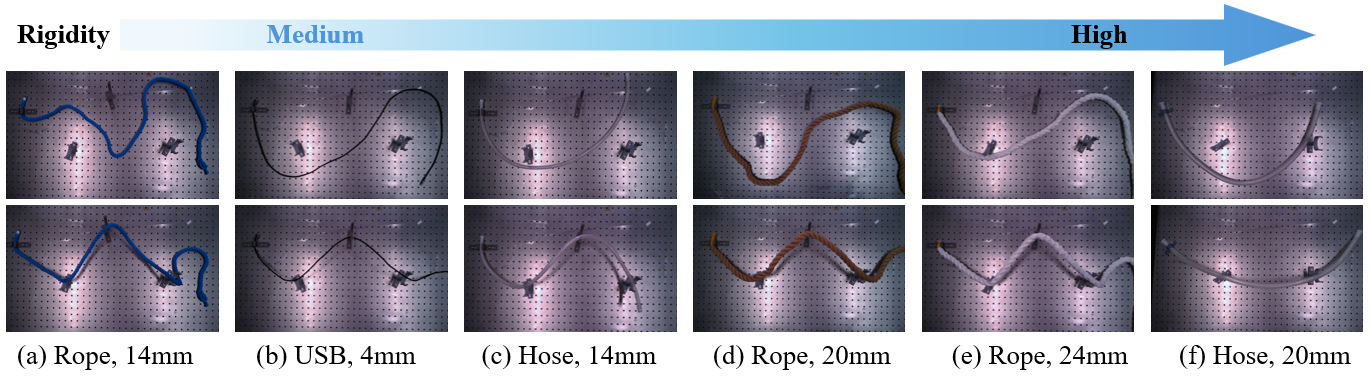}
    \caption{ Cable routing results with cable of different initial state, rigidities, diameters and random slots. First row: initial state. Second row: final state. }
    \label{routing}
\end{figure*}

\subsection{Grasping Test}
\label{sec5_2}

To estimate the proposed gripper fingernail's capability to grasp cable from a planar surface, we conduct a series of grasping experiments with some everyday DLOs like USB cables, hoses, and ropes, as shown in Fig. \ref{grasp}. 
The results show that our fingernails can not only accurately grasp cables of different diameters from the tabletop, but also maintain a small contact force and deformation during the grasping process, avoiding the extrusion damage caused by the traditional two-finger parallel gripper.

\subsection{Cable Routing Test}
To evaluate whether the designed framework can tackle cable routing tasks with varying cable and slot configurations, we conducted cable routing experiments with DLOs of different types, diameters, and rigidities. We randomly placed slots in different positions and orientations and randomized the initial state of the cables. Table \ref{tab:routing} shows that for cables of medium and large rigidities, our proposed framework performs well with a success rate of 31 out of 35 (Fig. \ref{routing}), and generalizes well to cables of different diameters. 

However, for DLOs with low rigidity, such as thin ropes, they are too soft and can cause stacking or folding, leading to task failure, as shown in Fig. \ref{cable_rout_fail} (b). In cases where the orientation of neighboring slots changes too much, the cable can also have difficulty aligning with slots and fail the task, as shown in Fig. \ref{cable_rout_fail} (c). Since most real cables have a certain rigidity (like the USB cable), our framework can perform well on real cable routing tasks in general. The folding and misalignment problems stem from limitations of single-arm operation, which can be solved after applying dual arms.



\section{Conclusion}
\label{sec5_3}

In this paper, we proposed a bio-inspired fingernail to help with cable grasping on planar surfaces without slippage and over-squeezing and conducted a series of
grasping experiments with some everyday linear objects to verify its capability to grasp various DLOs from the table.
Based on the mechanism of the fingernail, we introduced a single-grasp end-to-end 3D cable routing framework to complete cable routing tasks, reducing the perception-regrasp cycle of the normal pick-and-place strategy. Experiments demonstrated that this framework performs well (31/35) for cables
of medium and large rigidities  with random task configuration.

Our single-arm robotic manipulation framework still suffers from drawbacks in performing low-rigidity cable routing. In the future, we plan to apply a dual-arm robot for cable routing manipulation to solve the problem of cable folding and misalignment of low-rigidity cables.
More precise and safe cable manipulation can be achieved with multi-modal sensory information after installing tactile sensors on the fingernails, left as future directions to explore.


\begin{table}[t]
\centering
\caption{CABLE ROUTING WITH DIFFERENT CABLES}
\label{tab:routing}
{%
\small
\begin{tabular}{@{}lcccc@{}}
\toprule
Category & Dia./mm & Rigidity & Success rate & Failure reason\\
\midrule
\multirow{2}{*}{USB cable} 
& 4 & medium & 4/5 & A:0 B:1    \\
& 6 & medium & 5/5 & A:0 B:0    \\

\midrule

\multirow{2}{*}{PVC hose} 
& 14 & medium & 4/5 & A:0 B:1    \\
& 20 & high & 5/5 & A:0 B:0    \\

\midrule

\multirow{5}{*}{Nylon rope} 
& 4 & low & 0/5 & A:5 B:0    \\
& 8 & low & 1/5 & A:3 B:1    \\
& 14 & medium & 4/5 & A:1 B:0    \\
& 20 & high & 5/5 & A:0 B:0  \\
& 24 & high & 4/5 & A:0 B:1    \\
\bottomrule
\end{tabular}%
}
\footnotesize *A: Cable got folded. B: Cable are not aligned with slot.
\end{table}



\bibliographystyle{ieeetr} 
\bibliography{reference}

\begin{thebibliography}{10}

\bibitem{9732674}
C.~Wang, Y.~Zhang, X.~Zhang, Z.~Wu, X.~Zhu, S.~Jin, T.~Tang, and M.~Tomizuka, ``Offline-online learning of deformation model for cable manipulation with graph neural networks,'' {\em IEEE Robotics and Automation Letters}, vol.~7, no.~2, pp.~5544--5551, 2022.

\bibitem{9981313}
J.~Buzzatto, J.~Chapman, M.~Shahmohammadi, F.~Sanches, M.~Nejati, S.~Matsunaga, R.~Haraguchi, T.~Mariyama, B.~MacDonald, and M.~Liarokapis, ``On robotic manipulation of flexible flat cables: Employing a multi-modal gripper with dexterous tips, active nails, and a reconfigurable suction cup module,'' in {\em 2022 IEEE/RSJ International Conference on Intelligent Robots and Systems (IROS)}, pp.~1602--1608, 2022.

\bibitem{9571044}
I.~Armengol, A.~Suarez, G.~Heredia, and A.~Ollero, ``Design, integration and testing of compliant gripper for the installation of helical bird diverters on power lines,'' in {\em 2021 Aerial Robotic Systems Physically Interacting with the Environment (AIRPHARO)}, pp.~1--8, 2021.

\bibitem{chen2023contact}
K.~Chen, Z.~Bing, F.~Wu, Y.~Meng, A.~Kraft, S.~Haddadin, and A.~Knoll, ``Contact-aware shaping and maintenance of deformable linear objects with fixtures,'' in {\em 2023 IEEE/RSJ International Conference on Intelligent Robots and Systems (IROS)}, pp.~1--8, IEEE, 2023.

\bibitem{wilson2023cable}
A.~Wilson, H.~Jiang, W.~Lian, and W.~Yuan, ``Cable routing and assembly using tactile-driven motion primitives,'' in {\em 2023 IEEE International Conference on Robotics and Automation (ICRA)}, pp.~10408--10414, IEEE, 2023.

\bibitem{luo2307multi}
J.~Luo, C.~Xu, X.~Geng, G.~Feng, K.~Fang, L.~Tan, S.~Schaal, and S.~Levine, ``Multi-stage cable routing through hierarchical imitation learning. arxiv pre-print, 2023,'' {\em URL https://arxiv. org/abs/2307.08927}, vol.~22.

\bibitem{jin2022robotic}
S.~Jin, W.~Lian, C.~Wang, M.~Tomizuka, and S.~Schaal, ``Robotic cable routing with spatial representation,'' {\em IEEE Robotics and Automation Letters}, vol.~7, no.~2, pp.~5687--5694, 2022.

\bibitem{9347270}
Y.~Chen, Z.~Fang, S.~Liu, Y.~Wang, C.~Zhong, C.~Cai, Y.~Zhang, Y.~Wei, and Z.~Wang, ``A soft-robotic gripper for ultra-high-voltage transmission line operations,'' in {\em 2020 IEEE 4th Conference on Energy Internet and Energy System Integration (EI2)}, pp.~788--793, 2020.

\bibitem{10417506}
S.~Ishikawa, T.~Nishimura, and T.~Watanabe, ``Development of a gripper for manipulation of soft line-shaped object,'' in {\em 2024 IEEE/SICE International Symposium on System Integration (SII)}, pp.~225--226, 2024.

\bibitem{7139505}
E.~W. Hawkes, D.~L. Christensen, A.~K. Han, H.~Jiang, and M.~R. Cutkosky, ``Grasping without squeezing: Shear adhesion gripper with fibrillar thin film,'' in {\em 2015 IEEE International Conference on Robotics and Automation (ICRA)}, pp.~2305--2312, 2015.

\bibitem{Yu2024InHandFO}
M.~Yu, B.~Liang, X.~Zhang, X.~Zhu, X.~Li, and M.~Tomizuka, ``In-hand following of deformable linear objects using dexterous fingers with tactile sensing,'' {\em 2024 IEEE/RSJ International Conference on Intelligent Robots and Systems (IROS)}, pp.~13518--13524, 2024.

\bibitem{9816981}
P.~M. Fresnillo, S.~Vasudevan, and W.~M. Mohammed, ``An approach for the bimanual manipulation of a deformable linear object using a dual-arm industrial robot: cable routing use case,'' in {\em 2022 IEEE 5th International Conference on Industrial Cyber-Physical Systems (ICPS)}, pp.~1--8, 2022.

\bibitem{9197396}
J.~Zhao, X.~Wang, S.~Wang, X.~Jiang, and Y.~Liu, ``Assembly of randomly placed parts realized by using only one robot arm with a general parallel-jaw gripper,'' in {\em 2020 IEEE International Conference on Robotics and Automation (ICRA)}, pp.~5024--5030, 2020.

\bibitem{waltersson2022planning}
G.~A. Waltersson, R.~Laezza, and Y.~Karayiannidis, ``Planning and control for cable-routing with dual-arm robot,'' in {\em 2022 International Conference on Robotics and Automation (ICRA)}, pp.~1046--1052, IEEE, 2022.

\bibitem{suberkrub2022feel}
F.~S{\"u}berkr{\"u}b, R.~Laezza, and Y.~Karayiannidis, ``Feel the tension: Manipulation of deformable linear objects in environments with fixtures using force information,'' in {\em 2022 IEEE/RSJ International Conference on Intelligent Robots and Systems (IROS)}, pp.~11216--11222, IEEE, 2022.

\bibitem{de2018integration}
D.~De~Gregorio, R.~Zanella, G.~Palli, S.~Pirozzi, and C.~Melchiorri, ``Integration of robotic vision and tactile sensing for wire-terminal insertion tasks,'' {\em IEEE Transactions on Automation Science and Engineering}, vol.~16, no.~2, pp.~585--598, 2018.

\bibitem{10533619}
R.~Varghese and S.~M., ``Yolov8: A novel object detection algorithm with enhanced performance and robustness,'' in {\em 2024 International Conference on Advances in Data Engineering and Intelligent Computing Systems (ADICS)}, pp.~1--6, April 2024.

\bibitem{6630714}
J.~Schulman, A.~Lee, J.~Ho, and P.~Abbeel, ``Tracking deformable objects with point clouds,'' in {\em 2013 IEEE International Conference on Robotics and Automation}, pp.~1130--1137, 2013.

\bibitem{590011}
A.~Yuille and N.~Grzywacz, ``The motion coherence theory,'' in {\em [1988 Proceedings] Second International Conference on Computer Vision}, pp.~344--353, 1988.

\end{thebibliography}

\end{document}